\documentclass[runningheads]{llncs}
\usepackage{graphicx}
\usepackage{booktabs} 
\usepackage{hyperref}
\usepackage{comment}
\usepackage{multirow}
\begin{document}
\title{Fuzzy-Rough Nearest Neighbour Approaches for Emotion Detection in Tweets\thanks{This work was supported by the Odysseus programme of the Research Foundation—Flanders (FWO).}}
\titlerunning{FRNN Approaches for Emotion Detection in Tweets}
%
%
\author{Olha Kaminska \inst{1}\orcidID{0000-0002-4233-571X} 
\and Chris Cornelis
\inst{1}\orcidID{0000-0002-6852-4041} 
\and Veronique Hoste
\inst{2}\orcidID{0000-0002-0539-4630}
}
%
%
\institute{$^1$Computational Web Intelligence \\ Department of Applied Mathematics,  Computer Science and Statistics, \\ Ghent University, Ghent, Belgium \\
$^2$LT3 Language and Translation Technology Team \\ Ghent University, Ghent, Belgium \\
\email{\{Olha.Kaminska,Chris.Cornelis,Veronique.Hoste\}@UGent.be}}
\maketitle              
\begin{abstract}
Social media are an essential source of meaningful data that can be used in different tasks such as sentiment analysis and emotion recognition. Mostly, these tasks are solved with deep learning methods. Due to the fuzzy nature of textual data, we consider using classification methods based on fuzzy rough sets. \par 
Specifically, we develop an approach for the SemEval-2018 emotion detection task, based on the fuzzy rough nearest neighbour (FRNN) classifier enhanced with ordered weighted average (OWA) operators. We use tuned ensembles of FRNN--OWA models based on different text embedding methods. Our results are competitive with the best SemEval solutions based on more complicated deep learning methods. 

\end{abstract}

\section{Introduction}
\label{intro}

Over the past decades, the increasing availability of digital text material has allowed the domain of Natural Language Processing (NLP) to make significant headway in a wide number of applications, such as for example in the detection of hate speech \cite{macavaney2019hate} or emotion detection \cite{sailunaz2018emotion}.

In this paper, we report on our work on emotion detection for the SemEval-2018 Task 1 EI-oc: Affect in Tweets for English\footnote{\url{https://competitions.codalab.org/competitions/17751}} \cite{SemEval2018Task1}. This task represents a classification problem with tweets labeled with emotion intensity scores from 0 to 3 for four different emotions: anger, sadness, joy, and fear. \par 
We explored this task in our previous work \cite{kaminska-etal-2021-nearest}, using the weighted k Nearest Neighbours classification approach. We chose this method over popular neural network based solutions because of its explainability. Explainable models allow to investigate the classification progress and discover new patterns. \par 
Our purpose in this paper is to explore the efficiency of the fuzzy-rough nearest neighbour (FRNN) classifier \cite{jensen2011fuzzy} and its extensions based on ordered weighted average (OWA) operators \cite{cornelis2010ordered,lenz2019scalable} for this task. The motivation behind the usage of FRNN is to investigate the potential of relatively simple and transparent instance-based methods for the emotion detection task, in comparison with the black-box solutions offered by deep learning approaches. While the latter can solve sentiment analysis tasks with remarkable accuracy, they provide very little insight about how they reach their conclusions. This does not mean that we dismiss deep learning technology altogether; indeed, to prepare tweets for classification, we represent them by numerical vectors using some of the most popular current neural network based text embedding models \cite{barbieri2020tweeteval,cer2018universal,devlin2018bert,reimers2019sentence}. This strategy should allow us to strike the right balance between interpretability and accuracy of the approach.

\par 
The remainder of this paper has the following structure: Section \ref{works} contains an overview of related work, focusing on the SemEval-2018 Task 1 winning solutions. Section \ref{method} describes the main steps of our proposal, including data preprocessing and tweet representation and classification, and also recalls the competition's evaluation measures. Section \ref{experiments} reports on our approach's performance for the training data in different setups, while Section \ref{results} evaluates the best approach on the test data. Finally, Section \ref{conclusion} provides a discussion of the obtained results and some ideas for further research. 

The source code of this paper is available online at the GitHub repository\footnote{The source code: \url{https://github.com/olha-kaminska/frnn_emotion_detection}}.

\section{Related work}
\label{works}
We start this section by briefly describing the most successful solutions\footnote{Competition results: \url{https://competitions.codalab.org/competitions/17751\#results}} to the SemEval-2018 shared task. The winning approach \cite{duppada2018seernet} used ensembles of XGBoost and Random Forest classification models using tweet embedding vectors, while the second place was taken by \cite{gee2018psyml}, who used Long Short Term Memory (LSTM) neural nets with transfer learning. The third place contestants \cite{rozental2018amobee} presented a complex ensemble of models with Gated-Recurrent-Units (GRU) and a convolutional neural network (CNN) with the role of an attention mechanism. \par 
As is clear, the best approaches all used deep learning technology in one way or another, thus reflecting the current state-of-the-art and trends in automated text analysis (see e.g.\ \cite{minaee2020deep} for a comprehensive overview). This tendency is further reinforced by the use of the Pearson Correlation Coefficient (see formula (\ref{eq:pcc}) in Section 3.5) as the sole evaluation measure for the competition, since this measure lends itself well to NN-based optimization.\par 
To gain more insight into how tweets express different emotions and emotion intensities, instance-based methods may be used that discern tweets based on a similarity or distance metric. In particular, we want to explore the use of fuzzy rough set techniques for this purpose. We are not the first to do so: for example, in \cite{wang2015,wang2016approach}, Wang et al.\ used fuzzy rough set methods to discover emotions and their intensities in multi-label social media textual data.\par 
In this paper, we will use the fuzzy rough nearest neighbour (FRNN) classification algorithm originally proposed in \cite{jensen2011fuzzy}, and refined later with Ordered Weighted Average (OWA) operators \cite{cornelis2010ordered,lenz2019scalable}. 

\section{Methodology}
\label{method}

In this section, we describe the key ingredients of our methodology. At the data level, we first discuss the data preprocessing steps and then elaborate on the  different text embedding methods we implemented. Furthermore, we introduce the similarity relation we used to compare the tweet vectors and discuss the two main setups we used for classification, i.e., FRNN-OWA used as a standalone classifier and within an ensemble. We end the section by a description of the used evaluation method.\par 

The task we consider is the emotion intensity ordinal classification task (EI-oc, \cite{SemEval2018Task1}) for the emotions anger, fear, joy, and sadness. The aim is to classify an English tweet into one of four ordinal classes. Each class represents a level of emotion intensity: 0 stands for ``no emotion can be inferred", 1 corresponds to ``low amount of emotion can be inferred", 2 means ``moderate amount of emotion can be inferred", and 3 - ``high amount of emotion can be inferred". For each emotion, the training, development, and test datasets were provided in the framework of the SemEval-2018 competition. We merge training and development datasets for training our model.

\subsection{Data cleaning}

Before the embedding process, we may apply some operations to clean the tweets. In the first, general step, we delete account tags starting with $'@'$, extra white spaces, newline symbols ('$\backslash$n'), all numbers, and punctuation marks. We do not delete hashtags because they can be a source of useful information \cite{mohammad2015using}, so we just remove $'\#'$ symbols. Also, we replace '$\&$' with the word 'and' and replace emojis with their textual descriptions. We save emojis as they can be helpful for precision improvement \cite{wolny2016emotion}. Emojis are represented either by punctuation marks and/or a combination of letters, or as a small image decoded with Unicode. For the first type, we used their descriptions from the list of emoticons on Wikipedia\footnote{\url{https://en.wikipedia.org/wiki/List\_of\_emoticons}} for replacement. For the second type, we use the Python package \textit{``emoji"} \footnote{\url{https://pypi.org/project/emoji/}} for transformation.\par
The second step of tweet preprocessing is stop-word removal. For this purpose, the stop-words list from the NLTK package\footnote{\url{https://gist.github.com/sebleier/554280}} is used. \par 
Both general preprocessing and stop-word removal are optional for our purposes: during the experimental stage, we will examine whether they improve classification results or not. \par 
We also explored some important characteristics of the datasets and presented them in Table \ref{tab:owas}. One of characteristics is the class imbalance. It is quantified by the Imbalance Ratio (IR) which is equal to the ratio of the sizes of the largest and the smallest classes in the dataset.  \par

\begin{table}[!ht]
\centering
\caption{Characteristics of the combined train and development data for the four emotion datasets.}
\label{tab:owas}
\renewcommand{\arraystretch}{0.8}
\begin{tabular}{l|cccc}
	\toprule[1pt]\midrule[0.3pt]
	\bf{Characteristic} & \bf{Anger} & \bf{Joy} & \bf{Sadness} & \bf{Fear}\\
	\midrule\midrule
	IR & 1.677 & 1.47 & 2.2 & 8.04\\
	Size of the smallest class & 376 & 410 & 348 & 217 \\
	Number of instances & 2,089 & 1,906 & 1,930 & 2,641 \\
	\midrule[0.3pt]\bottomrule[1pt]
\end{tabular}
\end{table}

\subsection{Tweet embedding}

We represent each tweet as a vector, or set of vectors, to perform classification. For this purpose, we use the following word embedding techniques:
\begin{itemize}
    \item Gensim pre-trained Word2Vec\footnote{\url{https://drive.google.com/file/d/0B7XkCwpI5KDYNlNUTTlSS21pQmM}}, which contains a vocabulary with 3 million words and phrases and assigns a 300-dimension vector to each of them, obtained by training on a Google News dataset.  
    \item DeepMoji\footnote{\url{https://deepmoji.mit.edu/}} is a state-of-the-art sentiment embedding model. Millions of tweets with emojis were used to train the model to recognize emotions. DeepMoji provides for each sentence (tweet) a vector of size 2,304 dimensions. The model has implementations for several Python packages, and we used the one on PyTorch, made available by Huggingface\footnote{\url{https://github.com/huggingface/torchMoji}}. 
    \item Universal Sentence Encoder (USE) \cite{cer2018universal} is a sentence-level embedding method, which means it will create vectors for sentences or tweets as a whole. It was developed by the TensorFlow team\footnote{\url{https://www.tensorflow.org/hub/tutorials/semantic\_similarity\_with\_tf\_hub\_universal\_encoder}}. USE provides a 512-dimensional vector for a text paragraph (tweet), and was trained on several data sources for different NLP tasks such as text classification, sentence similarity, etc. The model was trained in two ways, using a deep averaging network (DAN) and a Transformer encoder. We chose the second type of USE after basic experiments for our further experiments.  
    \item Bidirectional Encoder Representations from Transformers (BERT), proposed by Devlin et al. \cite{devlin2018bert}. The Google AI Language Team developed a script\footnote{\url{https://github.com/dnanhkhoa/pytorch-pretrained-BERT/blob/master/examples/extract\_features.py}} that we use to assign pre-computed feature vectors with length 768 from a PyTorch BERT model to all the words of a tweet. If the BERT vocabulary does not contain some word, then during the embedding, this word is split into tokens (for example, if the word "$tokens$" is not in the BERT dictionary, then it can be represented as "$tok$", "$\#\#en$", "$\#\#s$"), and a vector is created for each token. 
    \item Sentence-BERT (SBERT) is a tuned and modified BERT model developed by Reimers et al.~\cite{reimers2019sentence}. The model operates on the sentence level and provides vectors with the same size as the original BERT. SBERT is based on siamese (twin) and triplet network structures, which can processes two sentences (tweets) simultaneously in the same way. 
    \item The Twitter-roBERTa-based model for Emotion Recognition presented by Barbieri et al.~\cite{barbieri2020tweeteval} provides embeddings on word level similar to the original BERT. We consider one of seven fine-tuned roBERTa-based models trained for different tasks with specific data for each of them. The model we chose was trained for the emotion detection task from the same SemEval competition (E-c) using a different set of tweets \cite{SemEval2018Task1} with emotions such as anger, joy, sadness, and optimism.
\end{itemize}
All listed sentence-level embeddings methods are applied to the tweets as a whole, while for the word- and token-level approaches, we calculated a tweet vector by taking its words' or tokens' vectors mean. The experiments were performed for all four emotion datasets and the obtained results are provided in Section \ref{experiments}.

\subsection{Similarity relation}

To be able to compare tweet vectors, we need an adequate similarity relation. We opted for the cosine metric, given by Formula~(\ref{eq:cos}): \cite{huang2008similarity}.
\begin{equation}
\label{eq:cos}
\cos(A, B) = \frac{A \cdot B}{||A|| \times ||B||},
\end{equation}
Here, $A$ and $B$ are elements from the same vector space, $A \cdot B$ is their scalar product, and $||x||$ is the vector norm of element $x$. \par 
As this metric returns values between -1 (perfectly dissimilar vectors), and 1 (perfectly similar vectors), we rescale them to [0,1] using Formula~(\ref{eq:cos_sim}) below, which we will use as our primary similarity relation.
\begin{equation}
\label{eq:cos_sim}
cos\_similarity(A, B) = \frac{1 + cos(A, B)}{2}.
\end{equation}

\subsection{Classification methods} 

In this section, we first recall the OWA-based Fuzzy  Rough  Nearest  Neighbor  (FRNN-OWA) classification method and then explain how to construct ensembles with it to solve the emotion detection task. \par 

\subsubsection{FRNN-OWA}
The fuzzy rough nearest neighbour (FRNN) method \cite{jensen2008new,jensen2011fuzzyb,jensen2011fuzzy} is an instance-based classifier that uses the lower (L) and upper (U) approximations from fuzzy rough set theory to make classifications. In order to make the method more robust and noise-tolerant, lower and upper approximations are usually calculated with Ordered Weighted Average (OWA) aggregation operators \cite{cornelis2010ordered}. The OWA aggregation of a set of values $V$ using weight vector $\overrightarrow{W} = \langle w_1,w_2,...,w_{|V|} \rangle$, with $(\forall i)(w_i \in [0, 1])$ and $\sum_{i=1}^{|V|} w_i = 1$, is given by Formula~(\ref{eq:owa}):

\begin{equation}
\label{eq:owa}
OWA_{\overrightarrow{W}}(V) = \sum_{i=1}^{|V|} (w_i v_{(i)}),
\end{equation}

where $v_{(i)}$ is the $i^{th}$ largest element in $V$.\par 
In this paper, we used the following 
types of OWA operators\footnote{$p$ refers to the number of elements in the OWA weight vector.}:
\begin{itemize}
    \item Strict weights, which contain only one non-zero position that does not depend on the actual values that are being aggregated:\\ 
    $\overrightarrow{W}_{L}^{strict}=\langle0,0,...,1\rangle$ $\overrightarrow{W}_{U}^{strict}=\langle1,0,...,0\rangle$.
    Strict weights correspond to the original FRNN proposal from \cite{jensen2008new}.
    \item Exponential weights (Exp), which are drawn from an exponential function with base 2: \\
    $\overrightarrow{W}_{L}^{exp} = \langle\frac{1}{2^p-1},\frac{2}{2^p-1},...,\frac{2^{p-2}}{2^p-1},\frac{2^{p-1}}{2^p-1}  \rangle$ \\ 
    $\overrightarrow{W}_{U}^{exp} = \langle \frac{2^{p-1}}{2^p-1},\frac{2^{p-2}}{2^p-1},...,\frac{2}{2^p-1},\frac{1}{2^p-1} \rangle$.
    \item Additive weights (Add), which model linearly decreasing or increasing weights: \\
    $\overrightarrow{W}_{L}^{add} = \langle\frac{2}{p(p+1)},\frac{4}{p(p+1)},...,\frac{2(p-1)}{p(p+1)},\frac{2}{p+1}  \rangle$ \\
    $\overrightarrow{W}_{U}^{add} = \langle \frac{2}{p+1},\frac{2(p-1)}{p(p+1)},...,\frac{4}{p(p+1)}, \frac{2}{p(p+1)} \rangle$.
    \item Inverse additive weights (Invadd) are also based on the ratio between consecutive elements in the weight vectors: \\
    $\overrightarrow{W}_{L}^{invadd} = \langle\frac{1}{pD_p},\frac{1}{(p-1)D_p},...,\frac{1}{2D_p},\frac{1}{D_p}  \rangle$ \\
    $\overrightarrow{W}_{U}^{invadd} = \langle \frac{1}{D_p},\frac{1}{2D_p},...,\frac{1}{(p-1)D_p}, \frac{1}{pD_p} \rangle$,\\
    with $D_p=\sum_{i=1}^{p} \frac{1}{p}$, the $p^{th}$ harmonic number.
    \item Mean weights, which weight each element equally:\\
$\overrightarrow{W}_{L}^{mean} = \overrightarrow{W}_{U}^{mean} = \langle\frac{1}{p},\frac{1}{p},...,\frac{1}{p}\rangle$
\end{itemize}
We used the implementation of the FRNN-OWA classifier \cite{lenz2019scalable} provided by the fuzzy-rough-learn package\footnote{https://github.com/oulenz/fuzzy-rough-learn}. To classify a test instance $y$, the method calculates its membership to the lower and upper approximation of each decision class $C$:
\begin{eqnarray}
\underline{C}(y) &=& OWA_{\overrightarrow{W}_{L}}\{1 - R(x,y) \mid x \in X \setminus C\}) \label{owalower} \\
\overline{C}(y) &=& OWA_{\overrightarrow{W}_{U}}\{R(x,y) \mid x \in C\}) \label{owaupper}
\end{eqnarray}
The algorithm then assigns $y$ to the class $C$ for which $\underline{C}(y) + \overline{C}(y)$ is highest.\par

Usually, the computation in Formula (\ref{owalower}) is restricted to the $k$ nearest neighbours of $y$ from the training data belonging to classes other than $C$, while in Formula (\ref{owaupper}) we consider only $y$'s $k$ nearest neighbours from class $C$.
There is no universal rule to determine the value of the parameter $k$. As a default, we can 
put $k =\frac{\sqrt{N}}{2}$, where $N$ is the size of the dataset. In order to examine the influence of $k$ on the obtained classification results, we will use different $k$ values for the best-performing approaches in our experiments for each dataset.
\par 
We performed experiments for each emotion dataset with different OWA types for lower and upper approximations with various numbers of $k$. \par 

\subsubsection{Classifier ensembles}
We used the FRNN-OWA method both as a standalone method and as part of a classification ensemble. For this purpose, a separate model was trained for every choice of tweet embedding. Each model was based on each dataset's best setup and embedding (choice of tweet preprocessing, OWA types, and the number of neighbours $k$). \par 
To determine the test label, we use a weighted voting function on the different outputs of our models. As possible voting functions $v$, we considered average, median, maximum, minimum, and majority. In the voting function the models' outputs receive some weights. \par
\begin{figure}[!ht]
\centering
  \includegraphics[width=1\linewidth]{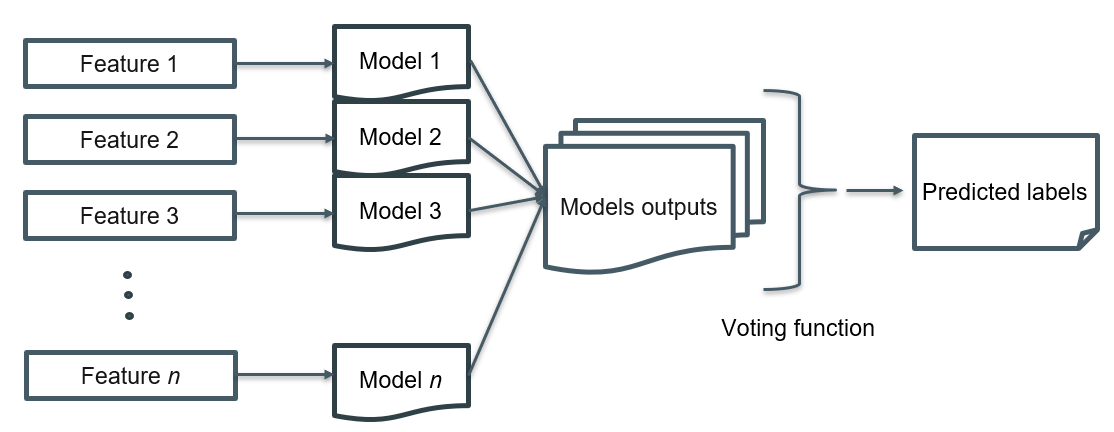}
  \caption{Scheme of the ensemble architecture.}
  \label{fig:ensemble}
\end{figure}

 
The full architecture of our ensemble approach is presented in Fig. \ref{fig:ensemble}. In Section \ref{experiments}, we perform several experiments to detect the most accurate ensemble setup, including the best voting function, the most suitable values of weights $\overrightarrow{E}$, and the proper combination of models (feature vectors).\par 

\subsection{Evaluation method}

We used 5-fold cross-validation to evaluate the results of our approaches. As evaluation measure the Pearson Correlation Coefficient (PCC) (\ref{eq:pcc}) was chosen, as it was also the evaluation measure used for the competition. \par
Assuming that $y$ is the vector of predicted values and $x$ is the vector of correct values, we compute

\begin{equation}
\label{eq:pcc}
PCC = \frac{\sum_i{({x_i}-\bar{x})({y_i}-\bar{y})}}{\sqrt{\sum_i{({x_i}-\bar{x})^2}\sum_i{({y_i}-\bar{y})^2}}},
\end{equation}

where ${x_i}$ and ${y_i}$ present the ${i}^{th}$  components of vectors $x$ and $y$ respectively and ${\bar{x}}$ and ${\bar{y}}$ are their means.\par
 The PCC measure provides a value between -1, which corresponds to a total negative linear correlation, and 1 - a total positive linear correlation, where 0 represents no linear correlation. Hence, the best classification model should provide the highest PCC. \par 
After submitting the obtained test labels to the competition web page, the PCC scores for each emotion dataset were averaged. \par 

\section{Experiments}
\label{experiments}

In this section, we present our results for the classification approaches discussed in the previous section. Initially, we explore the best individual FRNN-OWA setup, including the preprocessing options, the chosen tweet embedding, the  OWA types and the number of neighbours $k$. In a second set of experiments, we evaluate various ensemble approaches.

\subsection{Detecting the best setup for embeddings}

We performed experiments with different OWA types to detect the best setup for each dataset. We also investigated for each dataset whether it was beneficial to apply tweet preprocessing and stop-words cleaning. Finally, we explored the most suitable $k$ value for each embedding for each emotion dataset. 
\par 
First, the pipeline was performed for each embedding and emotion dataset after general preprocessing with the same OWA type for upper and lower approximations (strict, additive, exponential, and mean) and a different number of neighbors (from $5$ to $23$ with step $2$). As the results showed, the best results were obtained with the additive (``add'') OWA type for most embeddings, so we chose them for the further experiments. The best results for each dataset and each embedding are presented in Table~\ref{tab:embed}.\par 

\begin{table*}[!ht]
\centering
\caption{The best setup for each combination of dataset and embedding.}
\label{tab:embed}
\renewcommand{\arraystretch}{0.8}
\begin{tabular}{l|cccc}
    \toprule[1pt]\midrule[0.3pt]
    \bf{Setup} & \bf{Anger} & \bf{Joy} & \bf{Sadness} & \bf{Fear} \\
    \midrule\midrule
    \multicolumn{5}{c}{\bf{roBERTa-based}}\\
    Tweets preprocessing & Yes & Yes & Yes & Yes\\
    Stop-words cleaning & No & No & No & No\\
    Number of neighbors & 19&9&23&9\\ 
    PCC & 0.6779&0.6956&0.7062&0.6031\\
    \midrule
    \multicolumn{5}{c}{\bf{DeepMoji}}\\
    Tweets preprocessing & No & No & No & No\\
    Stop-words cleaning & No & No & No & No\\
    Number of neighbors & 23 & 19 & 23 & 21\\ 
    PCC & 0.5853 & 0.6520 & 0.6380 & 0.5745\\
    \midrule
    \multicolumn{5}{c}{\bf{BERT}}\\
    Tweets preprocessing & No & No & No & No \\
    Stop-words cleaning & No & No & No & No \\
    Number of neighbors & 19 & 17 & 23 & 7\\ 
    PCC & 0.4492 & 0.5374 & 0.4391 & 0.4500\\
    \midrule
    \multicolumn{5}{c}{\bf{SBERT}}\\
    Tweets preprocessing & Yes& Yes& Yes& Yes \\
    Stop-words cleaning & No & No & No & No\\
    Number of neighbors &19& 15&23&11\\ 
    PCC & 0.5016&0.5660&0.5655&0.5192\\
    \midrule
    \multicolumn{5}{c}{\bf{USE}}\\
    Tweets preprocessing &Yes & Yes & Yes & Yes\\
    Stop-words cleaning & No & No & No & No\\
    Number of neighbors & 23&23&23&21\\ 
    PCC & 0.5054&0.5693&0.5961&0.5764\\
    \midrule
    \multicolumn{5}{c}{\bf{Word2Vec}}\\
    Tweets preprocessing & Yes & Yes & Yes & Yes\\
    Stop-words cleaning & Yes & Yes & Yes & Yes \\
    The number of neighbors & 21 & 23 & 23 & 7\\ 
    PCC & 0.5009 & 0.5099 & 0.5048 & 0.4496\\
    \midrule[0.3pt]\bottomrule[1pt]
\end{tabular}
\end{table*}

Next, we calculated the PCC score for all embeddings and datasets with the best add OWA types, while varying the preparation level of the tweets: raw tweets (no preparation at all), standard preprocessing (text transformation steps mentioned in Section 3.1, excluding stop-words removal), and stop-words cleaning (the same as above, but including stop-words removal). To examine which setup works better, we performed a statistical analysis of results with a two-sided $t$-test (we assume the statistical significance of the $p$-value on the $0.05$ level). For calculation, the Python's package $'stats'$ was used. Results are presented in Table~\ref{tab:embed}. As we can see, some embeddings do not require any preprocessing at all, like DeepMoji and BERT. The standard preprocessing showed an improvement for other methods, and only Word2Vec seems to benefit from an additional stop-words removal step. \par 
For most of the experiments, the obtained $p$-values are below the chosen threshold of $0.05$. For some cases, the $p$-value was above the threshold, which means no significant difference exists between the compared options. In this situation, for the dataset, we chose the option that works better for other datasets. For example, for the BERT embedding, the joy dataset was the only one with the $p>0.05$, when for anger, sadness, and fear $p$ is below $0.05$ (so cleaned tweets performed better). Hence, we will use cleaned tweets for joy because this is the best setup for the other emotions.  \par 
Finally, for each embedding and dataset, we examined the PCC of the best setup (the combination of the best OWA types and the most efficient text preparation) for the different number of neighbours. The highest PCC scores and the proper $k$ values are also listed in Table~\ref{tab:embed}. \par 
The best setup for each combination of embedding and dataset was used in further experiments. We also can draw several intermediate conclusions. Remarkably, the highest PCC scores for all datasets among all embeddings were provided by the roBERTa-based model, which does not come as a surprise,since this model was fine-tuned on similar data and its performance is in line with earlier results for similar classification tasks \cite{barbieri2020tweeteval}. The second-best approach was DeepMoji, while BERT and Word2Vec provided the lowest scores. Also, we can see that the PCC scores for the fear dataset are often the lowest among the other emotions, which might probably be due to the fact that the fear dataset is the most unbalanced dataset. Similarly, the joy dataset shows high results, as the most balanced one.

\subsection{Ensembles}

To improve the PCC scores provided by individual embeddings, we also investigated an ensemble approach. To determine the best setup of the ensemble, we tuned several parameters, i.e., the voting function, the models' weights and the selection of the strongest embedding models. \par 
First, we compared different voting functions for all datasets: majority, mean, rounded mean, median, maximum, minimum. We note that for the majority voting function implementation we use the  $mode()$ function from the Python package $stats$. It chooses the most frequent label prediction, and in case of ties, this function returns the lowest value. \par 
Noteworthy is that some voting functions provide a float value between 0 and 3 instead of the required intensity labels 0, 1, 2, or 3. This was not a problem, though, during training because our labels are not different classes, but ordinal intensity labels. At testing time, the obtained values were rounded to submit our predictions. The general setup for comparing the voting functions was based on the six previously discussed models (one for each embedding method) with the parameters determined in Table~\ref{tab:embed}, where each predicted output has the same weight equal to $1$. The results are presented in Table~\ref{tab:voting}. As we can see, the mean voting function consistently provided the best results for all datasets, while median performs second best. Although the rounding of the mean's output decreases the PCC results, it remains the best voting function. So, for further experiments, we will use the average as a voting function. \par 

\begin{table*}[!ht]
\centering
\caption{Results for ensembles with different voting functions for all datasets.}
\label{tab:voting}
\renewcommand{\arraystretch}{0.8}
\begin{tabular}{l|cccc}
    \toprule[1pt]\midrule[0.3pt]
\bf{Voting function} & \bf{Anger} & \bf{Joy} & \bf{Sadness} & \bf{Fear}\\
    \midrule\midrule
    Majority&0.6141&0.6669&0.6591&0.5665\\ 
    Mean&\bf{0.6933}&\bf{0.7501}&\bf{0.7456}&\bf{0.6723}\\
    Rounded mean&0.6485&0.7126&0.7152&0.6448\\
    Median&0.6414&0.7150&0.7079&0.6050\\
    Maximum&0.4856&0.4668&0.5625&0.5640\\
    Minimum&0.5959&0.6411&0.5016&0.3885\\
    \midrule[0.3pt]\bottomrule[1pt]
\end{tabular}
\end{table*}

Next, we check the use of weights assigned to the models' outputs in the voting function. In particular, we use confidence scores (CS) to give more weight to the better models.\par 
A confidence score is a float value, usually between $0$ and $1$, provided by a classification model for each prediction class. This value illustrates the accuracy of the model's prediction for a particular class. For FRNN-OWA, the models return four scores (one for each class). They are the mean membership degrees in the upper and lower approximations. 
\par
To get confidence scores, we divide each score by the sum of all four class scores. In this way, we obtain the values $C_{i,j}$: four scores (one per class label $i, i = 0,...,3$) for every model $j$  ($j=1,...,6$). We 
use the confidence scores in the following ways:
\begin{itemize}
    \item Majority voting. The most intuitive approach, where we take as a prediction the label with the highest sum of confidence scores.
    \item Weighted average (WA). As we saw above, the best voting function is the mean, so we will upgrade it with confidence scores as weights to calculate the prediction label as a weighted average of labels. The output could be a float number, so we also check the rounded option.
\end{itemize}
Experiments were performed with all six embedding models. Results are provided in the upper half of Table~\ref{tab:conf}. \par 
\begin{table*}[!ht]
\centering
\caption{Results for ensembles with different usage of confidence scores for all datasets.}
\label{tab:conf}
\renewcommand{\arraystretch}{0.8}
\begin{tabular}{l|cccc}
    \toprule[1pt]\midrule[0.3pt]
\bf{Approach} & \bf{Anger} & \bf{Joy} & \bf{Sadness} & \bf{Fear}\\
    \midrule\midrule
    \multicolumn{5}{c}{Original confidence scores}\\
    \midrule
    Majority voting& 0.6351&0.7082&0.7016&0.5700\\
    Weighted average& 0.7025&0.7424&0.7333&0.6044\\
    WA rounded& 0.6302&0.6731&0.6962&0.5549\\
    \midrule\midrule
    \multicolumn{5}{c}{Rescaled confidence scores}\\
    \midrule
    Weighted average&\bf{0.7187}&\bf{0.7781}&\bf{0.7630}&\bf{0.6763}\\
   & ($\alpha=0.0420$) & ($\alpha=0.0360$) & ($\alpha=0.0400$)&($\alpha=0.0460$) \\
   
    WA rounded& 0.6432&0.7512& 0.7455& 0.6430 \\
   & ($\alpha=0.0420$) & ($\alpha=0.0320$) & ($\alpha=0.0320$)& ($\alpha=0.0460$) \\
    \midrule[0.3pt]\bottomrule[1pt]
\end{tabular}
\end{table*}
As we can see, weighted average with confidence scores performed the best. Predictably, rounding decreased the weighted average's score, and it is similar to the results provided by a majority of confidence scores. If we compare them with the values in Table~\ref{tab:voting}, considering the mandatory rounding step, we can conclude that these approaches with confidence scores do not increase PCC scores. \par
We analyzed the obtained confidence scores and noticed that they are close to each other, approximately, in the range from $0.4$ to $0.6$. Our hypothesis is that since we have a high dimensional task like ours, the confidence scores will be close to $0.5$: the upper approximation memberships will be close to $1$ and the lower ones to $0$, resulting in similar values for each class. In other words, the contribution of such a classifier is low.  \par

To mend this issue, we perform rescaling of the original membership scores in order to increase the differences among them. For this purpose, we subtract the mean $0.5$ from each score $C_{i,j}$ and divide the result by a small value $\alpha$ ($0<\alpha<1$). Next, for each class $i$ we compute the sum of the scores for each model. Since the obtained values may be negative, we use the softmax transformation to turn them into probabilities. The steps of this rescaling process are summarized in Formula (\ref{eq:rescale}):
\begin{equation}
\label{eq:rescale}
C_{i} = \frac{\exp(\sum_j (C_{i,j}-0.5) /\alpha)}{\sum_k \exp(\sum_j (C_{k,j}-0.5) /\alpha)} , 
\end{equation}
where 
$\alpha$ is a parameter to tune. To detect the best value of $\alpha$ for each dataset, we performed a grid search, calculating PCC scores for different $\alpha$ values to choose the one that provides the biggest PCC. \par 
Finally, to calculate the predicted label, we apply the weighted average on classes, where weights are calculated probabilities. 
Results of this approach with the best $\alpha$ for each dataset are provided in the lower half of Table~\ref{tab:conf}. \par
Compared with the original confidence scores and values from Table~\ref{tab:voting}, scaled scores performed better for each dataset for both average and rounded average. Hence, we will use scaled confidence scores as models' output weights in the following experiments. \par
The last step of ensemble tuning is to determine the most accurate set of models in the ensemble. The idea behind this is to see how the PCC score will change depending on the models (embeddings) that we are using in the ensemble to answer the question: is it possible to improve the score by rejecting the weak models' results.\par 
For this purpose, we used grid search, where the PCC score was calculated for each subset of all six models (features) and compared. The predicted label was calculated using a rounded average function with weights equal to the scaled confidence scores. We used a rounded average since it returns integers, so we can use them to submit to the competition web-page. In this way, we detected the best setup for each emotion dataset. The results for cross-validation evaluation are presented in Table~\ref{tab:test}. \par 
As we can see, all datasets have in common the same features such as roBERTa, DeepMoji, and USE models (we denote them with ``r/D/U''). Another one or two features are different for each dataset. We can mainly see that more features provide better results, but the weak models' pruning also takes place. \par 
In the end, we could obtain the best ensemble setup with the required parameters for each emotion dataset. 

\section{Results on the test data}
\label{results}

From Section \ref{experiments} we obtained the best setup for each dataset: an ensemble of several models based on different features with proper text preprocessing, $k$ value, and additive lower and upper OWA types for each. The predicted test label is calculated as the mean of the models' outputs with scaled confidence scores as weights.\par 
To measure the best ensemble's effectiveness, we evaluate it on the test data. We calculate PCC values for each emotion dataset and average the results, as was done by the competition organizers. As the output of the ensemble's mean voting function, obtained predictions are in float format, so to satisfy the competition's submitting format, they were rounded to the nearest integer value. The obtained results are presented in Table~\ref{tab:test}, where we  provided results for the combined training and development data to compare them. \par

\begin{table}[!ht]
\centering
\caption{The best approach on the cross-validation and test data for all datasets.}
\label{tab:test}
\renewcommand{\arraystretch}{0.8}
\begin{tabular}{l|l|cc}
    \toprule[1pt]\midrule[0.3pt] 
\bf{Dataset} & \bf{Models}&\bf{Training and} & \bf{Test data}
\\
&& \bf{development data} &\\
    \midrule\midrule
    Anger& r/D/U, Word2Vec, BERT & 0.7241 & 0.6388\\
    Joy& r/D/U, SBERT, BERT & 0.7788 &0.7115\\
    Sadness& r/D/U, SBERT & 0.7719& 0.6967 \\
    Fear& r/D/U, Word2Vec, SBERT &0.6930&0.5705 \\
    \midrule
    Averaged scores&&0.7419&0.6544\\
    \midrule[0.3pt]\bottomrule[1pt]
\end{tabular}
\end{table}

As we can see from Table~\ref{tab:test}, results for the test data are predictably worse than those for the combined training and development datasets. The PCC scores for sadness and joy datasets are higher than for anger, and fear, as usual, has lower results.
\par 
We submitted the predicted labels for the test data in the required format to the competition webpage\footnote{\url{https://competitions.codalab.org/competitions/17751\#learn_the_details-evaluation}}. After submission, we took the second place in the competition leader board with PCC = $0.654$. 

\section{Conclusion and future work}
\label{conclusion}
In this paper, we designed a weighted ensemble of FRNN-OWA classifiers to tackle the emotion detection task. Our approach uses several embeddings, which are mostly sentiment-oriented and applied at sentence-level. We demonstrated that our method, despite its simple design, is competitive to the competition's winning approaches, which are all black-boxes.  \par 
As a possible improvement, we may consider additional text preparation steps, for example, bigger weights for hashtags and emojis or exclamation mark usage, before the embedding step. \par 
Finally, we hypothesize that the lower PCC scores for the fear dataset could be related to the dataset's imbalance. As a possible approach to solve this issue, we may use specific classification machine learning methods for imbalanced data. For example, in paper \cite{vluymans2019dealing}, several fuzzy rough set theory methods are described specifically targeting imbalanced data sets.  

\bibliographystyle{splncs04}
\bibliography{samplepaper}

\begin{thebibliography}{10}
\providecommand{\url}[1]{\texttt{#1}}
\providecommand{\urlprefix}{URL }
\providecommand{\doi}[1]{https://doi.org/#1}

\bibitem{barbieri2020tweeteval}
Barbieri, F., Camacho-Collados, J., Espinosa~Anke, L., Neves, L.:
  {T}weet{E}val: Unified benchmark and comparative evaluation for tweet
  classification. In: Findings of the Association for Computational
  Linguistics: EMNLP 2020. pp. 1644--1650. Association for Computational
  Linguistics, Online (Nov 2020). \doi{10.18653/v1/2020.findings-emnlp.148},
  \url{https://www.aclweb.org/anthology/2020.findings-emnlp.148}

\bibitem{cer2018universal}
Cer, D., Yang, Y., Kong, S.y., Hua, N., Limtiaco, N., St.~John, R., Constant,
  N., Guajardo-Cespedes, M., Yuan, S., Tar, C., Strope, B., Kurzweil, R.:
  Universal sentence encoder for {E}nglish. In: Proceedings of the 2018
  Conference on Empirical Methods in Natural Language Processing: System
  Demonstrations. pp. 169--174. Association for Computational Linguistics,
  Brussels, Belgium (Nov 2018). \doi{10.18653/v1/D18-2029},
  \url{https://www.aclweb.org/anthology/D18-2029}

\bibitem{cornelis2010ordered}
Cornelis, C., Verbiest, N., Jensen, R.: Ordered weighted average based fuzzy
  rough sets. In: International Conference on Rough Sets and Knowledge
  Technology. pp. 78--85. Springer (2010)

\bibitem{devlin2018bert}
Devlin, J., Chang, M.W., Lee, K., Toutanova, K.: Bert: Pre-training of deep
  bidirectional transformers for language understanding. In: Proceedings of the
  2019 Conference of the North American Chapter of the Association for
  Computational Linguistics: Human Language Technologies, Volume 1 (Long and
  Short Papers). pp. 4171--4186 (2019)

\bibitem{duppada2018seernet}
Duppada, V., Jain, R., Hiray, S.: {S}eer{N}et at {S}em{E}val-2018 task 1:
  Domain adaptation for affect in tweets. In: Proceedings of The 12th
  International Workshop on Semantic Evaluation. pp. 18--23. Association for
  Computational Linguistics, New Orleans, Louisiana (Jun 2018).
  \doi{10.18653/v1/S18-1002}, \url{https://www.aclweb.org/anthology/S18-1002}

\bibitem{gee2018psyml}
Gee, G., Wang, E.: psyml at semeval-2018 task 1: Transfer learning for
  sentiment and emotion analysis. In: Proceedings of The 12th International
  Workshop on Semantic Evaluation. pp. 369--376 (2018)

\bibitem{huang2008similarity}
Huang, A.: Similarity measures for text document clustering. In: Proceedings of
  the sixth new zealand computer science research student conference
  (NZCSRSC2008), Christchurch, New Zealand. vol.~4, pp. 9--56 (2008)

\bibitem{jensen2008new}
Jensen, R., Cornelis, C.: A new approach to fuzzy-rough nearest neighbour
  classification. In: International conference on rough sets and current trends
  in computing. pp. 310--319. Springer (2008)

\bibitem{jensen2011fuzzyb}
Jensen, R., Cornelis, C.: Fuzzy-rough nearest neighbour classification. In:
  Transactions on rough sets XIII, pp. 56--72. Springer (2011)

\bibitem{jensen2011fuzzy}
Jensen, R., Cornelis, C.: Fuzzy-rough nearest neighbour classification and
  prediction. Theoretical Computer Science  \textbf{412}(42),  5871--5884
  (2011)

\bibitem{kaminska-etal-2021-nearest}
Kaminska, O., Cornelis, C., Hoste, V.: Nearest neighbour approaches for emotion
  detection in tweets. In: Proceedings of the Eleventh Workshop on
  Computational Approaches to Subjectivity, Sentiment and Social Media
  Analysis. pp. 203--212. Association for Computational Linguistics, Online
  (Apr 2021), \url{https://www.aclweb.org/anthology/2021.wassa-1.22}

\bibitem{lenz2019scalable}
Lenz, O.U., Peralta, D., Cornelis, C.: Scalable approximate frnn-owa
  classification. IEEE Transactions on Fuzzy Systems  \textbf{28}(5),  929--938
  (2019)

\bibitem{macavaney2019hate}
MacAvaney, S., Yao, H.R., Yang, E., Russell, K., Goharian, N., Frieder, O.:
  Hate speech detection: Challenges and solutions. PloS one  \textbf{14}(8),
  e0221152 (2019)

\bibitem{minaee2020deep}
Minaee, S., Kalchbrenner, N., Cambria, E., Nikzad, N., Chenaghlu, M., Gao, J.:
  Deep learning based text classification: A comprehensive review. arXiv
  e-prints pp. arXiv--2004 (2020)

\bibitem{SemEval2018Task1}
Mohammad, S.M., Bravo-Marquez, F., Salameh, M., Kiritchenko, S.: Semeval-2018
  {T}ask 1: {A}ffect in tweets. In: Proceedings of International Workshop on
  Semantic Evaluation (SemEval-2018). New Orleans, LA, USA (2018)

\bibitem{mohammad2015using}
Mohammad, S.M., Kiritchenko, S.: Using hashtags to capture fine emotion
  categories from tweets. Computational Intelligence  \textbf{31}(2),  301--326
  (2015)

\bibitem{reimers2019sentence}
Reimers, N., Gurevych, I.: Sentence-{BERT}: Sentence embeddings using {S}iamese
  {BERT}-networks. In: Proceedings of the 2019 Conference on Empirical Methods
  in Natural Language Processing and the 9th International Joint Conference on
  Natural Language Processing (EMNLP-IJCNLP). pp. 3982--3992. Association for
  Computational Linguistics, Hong Kong, China (Nov 2019).
  \doi{10.18653/v1/D19-1410}, \url{https://www.aclweb.org/anthology/D19-1410}

\bibitem{rozental2018amobee}
Rozental, A., Fleischer, D.: {A}mobee at {S}em{E}val-2018 task 1: {GRU} neural
  network with a {CNN} attention mechanism for sentiment classification. In:
  Proceedings of The 12th International Workshop on Semantic Evaluation. pp.
  218--225. Association for Computational Linguistics, New Orleans, Louisiana
  (Jun 2018). \doi{10.18653/v1/S18-1033},
  \url{https://www.aclweb.org/anthology/S18-1033}

\bibitem{sailunaz2018emotion}
Sailunaz, K., Dhaliwal, M., Rokne, J., Alhajj, R.: Emotion detection from text
  and speech: a survey. Social Network Analysis and Mining  \textbf{8}(1),
  1--26 (2018)

\bibitem{vluymans2019dealing}
Vluymans, S.: Dealing with imbalanced and weakly labelled data in machine
  learning using fuzzy and rough set methods. Springer (2019)

\bibitem{wang2015}
Wang, C., Feng, S., Wang, D., Zhang, Y.: Fuzzy-rough set based multi-labeled
  emotion intensity analysis for sentence, paragraph and document. In: Li, J.,
  Ji, H., Zhao, D., Feng, Y. (eds.) Natural Language Processing and Chinese
  Computing. pp. 444--452. Springer International Publishing, Cham (2015)

\bibitem{wang2016approach}
Wang, C., Wang, D., Feng, S., Zhang, Y.: An approach of fuzzy relation equation
  and fuzzy-rough set for multi-label emotion intensity analysis. In:
  International Conference on Database Systems for Advanced Applications. pp.
  65--80. Springer (2016)

\bibitem{wolny2016emotion}
Wolny, W.: Emotion analysis of twitter data that use emoticons and emoji
  ideograms  (2016)

\end{thebibliography}

\end{document}